\RequirePackage[T1]{fontenc} 
\documentclass[journal]{IEEEtran}
\usepackage{amsmath,amsfonts}
\usepackage{algorithmic}
\usepackage{algorithm}
\usepackage{array}
\usepackage[caption=false,font=normalsize,labelfont=sf,textfont=sf]{subfig}
\usepackage{textcomp}
\usepackage{stfloats}
\usepackage{url}
\usepackage{verbatim}
\usepackage{graphicx}
\usepackage{cite}
\usepackage{multirow} 
\usepackage{booktabs} 
\usepackage{bm} 
\usepackage{hyperref} 
\usepackage[capitalise]{cleveref} 

\begin{document}

\title{Rethinking the initialization of Momentum in Federated Learning with Heterogeneous Data}

\author{Chenguang Xiao, Shuo Wang
  \thanks{C. Xiao and S. Wang are with School of Computer Science, University of Birmingham, Birmingham, UK.}
}

\markboth{}%
{Shell \MakeLowercase{\textit{et al.}}: A Sample Article Using IEEEtran.cls for IEEE Journals}


\maketitle

\begin{abstract}
  Data Heterogeneity is a major challenge of Federated Learning performance.
  Recently, momentum based optimization techniques have beed proved to be effective in mitigating the heterogeneity issue.
  Along with the model updates, the momentum updates are transmitted to the server side and aggregated.
  Therefore, the local training initialized with a global momentum is guided by the global history of the gradients.
  However, we spot a problem in the traditional cumulation of the momentum which is suboptimal in the Federated Learning systems.
  The momentum used to weight less on the historical gradients and more on the recent gradients.
  This however, will engage more biased local gradients in the end of the local training.
  In this work, we propose a new way to calculate the estimated momentum used in local initialization.
  The proposed method is named as Reversed Momentum Federated Learning (RMFL).
  The key idea is to assign exponentially decayed weights to the gradients with the time going forward, which is on the contrary to the traditional momentum cumulation.
  The effectiveness of RMFL is evaluated on three popular benchmark datasets with different heterogeneity levels.

\end{abstract}


\section{Introduction}
\IEEEPARstart{F}{ederated} learning is designed as a loop of two stage training process when initially proposed~\cite{BrendanMcMahan2017}.
The first stage is the local training on the client side with the concurrent global model.
The second stage is the aggregation of the local model updates on the server side.

This special design isolates the central server from the client data, making the collective learning without sharing the raw data possible.
However, the two stage training process differs Federated Learning (FL) from the centralized training and introduces new challenges.
One of the major challenges is the data heterogeneity among the clients.

We refer the data heterogeneity as the difference of the data distribution among the clients.
This is also known as the Non-IID data distribution.
Sometimes, it can be described as the class imbalance discrepancy between clients.

With heterogeneous data in FL, the objective function of each client becomes different.
This difference increases along with the level of heterogeneity and makes the client model updates biased.
The server aggregating stage is largely affected by the biased model updates as weighted average of biased updates is a poor approximation of the global objective function.
Consequently, longer iterations and more communication rounds are required for the convergence of the global model with even worse accuracy.

Advanced optimization techniques are widely adopted in centralized training, and show great potential in FL heterogeneity issue.
Two main advance in centralized optimization are the momentum and adaptive learning rate methods.
Sometimes, they are combined to achieve better performance.
The naive practice of applying the advanced optimization techniques in FL is to treat each local training isolated.
These methods are direct but suboptimal as the momentum are initialized to zero in each local training.
This breaks the continuity of the momentum in different global iterations.

There are some works trying to bridge the gap between the centralized optimization and the FL optimization.
MFL~\cite{liu2020accelerating} is one of the most popular momentum based optimization techniques in FL systems.
Along with the model updates, MFL transmits the momentum updates to the server side and aggregates them with the model updates.
The new global model and the aggregated momentum are broadcasted to the clients to initiate the next round of local training.

Similarly, Adam can be applied in FL systems with the same idea.
Aside from the model updates, the server aggregates the first and second moment estimates of the gradients.
The aggregated estimates are broadcasted to the clients to initialize the next round of local training.
The first moment works by applying momentum to the gradients and the second moment works by adapting the learning rate for each parameter.

Those methods, however, still confront the heterogeneity issue when initializing the momentum.
Briefly, the less accurate gradient at the end of local training count for more in the cumulative momentum.
This makes the aggregated momentum biased and suboptimal for the local training initialization.

In this work, we propose a new way to calculate the estimated momentum used in local initialization.

\section{Background}
As a two stage training process, FL optimization are in two levels.
The local training on the clients is exactly the same as the centralized training.
The global training can be viewed as a Stochastic Gradient Descent (SGD) step where the gradients are the average of the local gradients.

\subsection{Advanced Optimization Techniques}
SGD is the most popular optimization technique in machine learning with computational efficiency.
Based on SGD, there are two major advanced optimization techniques: momentum and adaptive learning rate methods.

\begin{figure}[htbp]
  \centering
  \includegraphics{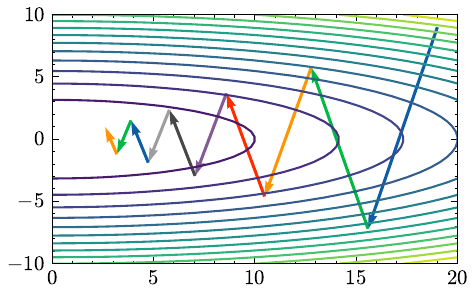}
  \caption{Stochastic Gradient Descent}
  \label{fig:sgd}
\end{figure}

\begin{figure}[htbp]
  \centering
  \includegraphics{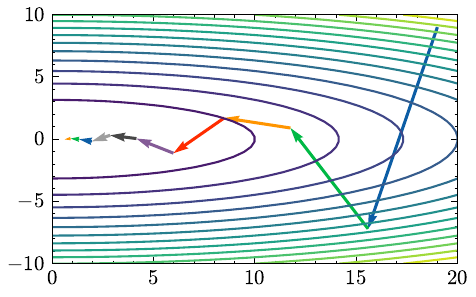}
  \caption{Stochastic Gradient Descent with Momentum}
  \label{fig:sgdm}
\end{figure}

The momentum methods use an exponential moving average of all past gradients to update the model instead of the current gradient solely.
This modification makes the descent step keeps a balance between gradient at current point and its neighbourhood.
With proper momentum coefficient, the momentum methods can accelerate the convergence and avoid the oscillation in the optimization process.
Intuitively, the momentum methods act as a heavy ball rolling down the hill as show in \cref{fig:sgd,fig:sgdm}.

Learning rate plays a crucial role in the optimization process.
Larger learning rate benefits the initial training process and smaller learning rate benefits the training close to the optimum.
Asides from the manual learning rate scheduling, the adaptive learning rate methods adjust the learning rate automatically.
Adam is one of the most popular adaptive learning rate methods which employs the second moment of gradient to adjust the learning rate for each parameter.

In this work, we focus on exploring the best practice of applying momentum in FL systems.

\subsection{Federated Optimization}
Federated Averaging (FedAvg) is the most popular optimization technique in FL systems.
The local training of FedAvg is simply the SGD while the global training is the aggregation of the local model updates.

The global step of FedAvg can also be viewed as a naive SGD step where the gradients are the average of the local gradients.
Advanced optimization techniques benefits the aggregation especially when the data distribution among the clients are heterogeneous.

MFL~\cite{liu2020accelerating} applies SGD with momentum in the local training and aggregates the momentum buffer in the server side.
With the aggregated momentum buffer broadcasted to the clients for the initialization of the next round of local training, MFL shows faster convergence and better accuracy in heterogeneity.

SCAFFOLD~\cite{DBLP:conf/icml/KarimireddyKMRS20} leverages the global and local control variable to correct the bias in the local model updates.

FedDyn~\cite{DBLP:conf/icpads/JinCGL22} uses a dynamic regularization term to adjust the local model updates.
The dynamic regularization term is calculated based on the local model updates and the global model.

FedOpt~\cite{DBLP:conf/iclr/ReddiCZGRKKM21} is a series of optimization techniques used in the server aggregation stage.

FedCM~\cite{xu2021fedcm} add a constant momentum to the local model updates, which forms a global momentum in the server aggregation stage.

FAFED~\cite{DBLP:conf/aaai/WuHHH23} is a faster adaptive federated learning method which uses a dynamic learning rate and a dynamic momentum coefficient.

FedAMS~\cite{DBLP:conf/icml/WangLC22} is a federated learning method with adaptive momentum scaling.


Despite the success of momentum based optimization and adaptive learning rate techniques in FL systems, there is still room for improvement, especially in the data heterogeneity.
One of the major challenges is the calculation of the estimated momentum used in local initialization.
\section{Problem Statement}
The aggregated momentum in MFL fails to capture the gradient at the neighbourhood of the global model with heterogeneous data distribution.
Traditional momentum cumulation assigns more weights to the recent gradients than the past gradients.
However, the recent gradients are biased due to the heterogeneity among the clients.
Using centralized momentum cumulation in FL could be suboptimal.
In this section, we will show you the problem and propose a solution accordingly.

The Momentum cumulation is accurate only if all the client data are IID.
The data heterogeneity results in the diverged gradients among the clients.
And the bias increase with the level of heterogeneity among the clients and the steps of local training.

\begin{figure}[htbp]
  \centering
  \includegraphics{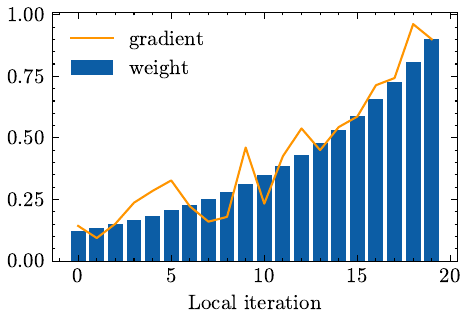}
  \caption{Illustration of the momentum cumulation in FL systems.}
  \label{fig:mom-illust}
\end{figure}

\Cref{fig:mom-illust} illustrates the momentum in a client with heterogeneous data distribution.
The blue bar indicates the weights of the gradients at given local iterations.
The weight increases exponentially during the training process.
The orange curve indicates the gradient bias from the global model.
With longer local training, the gradient bias increases gradually as a result of the heterogeneity.
Obviously, weighted sum of the gradients in \cref{fig:mom-illust} address more basis in the later local training.
In this way, the biased gradients dominate the cumulative momentum and mislead the next local momentum initialization.

\subsection{Case Study}
We use a simple case study to illustrate the problem of current momentum cumulation in FL systems.
Recall the momentum update rule in SGD with momentum:
\begin{equation}
  \label{eq:momentum}
  \begin{aligned}
    v_t & = \beta v_{t-1} + \nabla f(x_t)                          \\
        & = \beta^t v_0 + \sum_{i=1}^{t} \beta^{t-i} \nabla f(x_i)
  \end{aligned}
\end{equation}
where $v_t$ is the momentum at time $t$, $\beta$ is the momentum coefficient, $f(x_t)$ is the gradient of the loss function at time $t$.
The momentum update rule keeps the same in local training of FL systems.

However, the local training in different global iterations are not directly linked as in centralized training.
The server aggregates the local model updates and the momentum values for the initialization of the next round of local training.
This is a crucial difference between FL and centralized training and causes a well known issue of the FL systems.
We call this issue as sensitivity to the number of local epochs.

The longer the local training in single global iteration, the more biased the model updates are.
This holds true for the cumulative momentum as well.
As the cumulative momentum are calculated as the Exponential Moving Average (EMA) of the gradients, more weights are put on the recent gradients than the past gradients.
Combining the truth that the bias increases gradually in the training process, the cumulative momentum will contain more bias in the later local training.

\begin{figure}[htbp]
  \centering
  \includegraphics{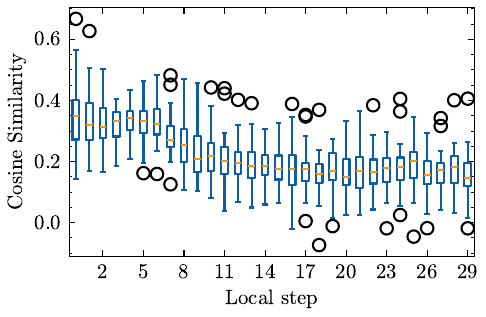}
  \caption{Box plot of the cumulative momentum in different global iterations.}
  \label{fig:cs-box}
\end{figure}

The gradient divergence between heterogeneous clients can be formulated as:
\begin{equation}
  \label{eq:divergence}
  \nabla f^{(i)}_{t,k}(x_{k-1}^{(i)}) - \nabla f^{(j)}_{t,k}(x_{k-1}^{(j)}) \leq \epsilon_k
\end{equation}
where $\nabla f^{(i)}(x_t)$ is the gradient of the loss function at global iteration $t$ and local step $k$ on client $i$, $\epsilon_k$ is the divergence upper bound.
The expectation of the upper bound $\epsilon_k$ is increasing with the number of local epochs.
Combined with \cref{eq:momentum}, the cumulative momentum weights more on the latest diverged gradients than the past one.
As shown \cref{fig:cs-box}, the cumulative momentum in the later global iterations are more biased than the earlier ones.

We design an experiment to show the gradients' divergence in heterogeneous FL systems with CIFAR10 dataset.
The experiment runs 100 global iterations with 10 clients in each round.
In a single local training, the clients perform 30 SGD steps.
The gradients for each client at each local step are logged to show the divergence in the later global iterations.

\begin{figure}[htbp]
  \centering
  \includegraphics{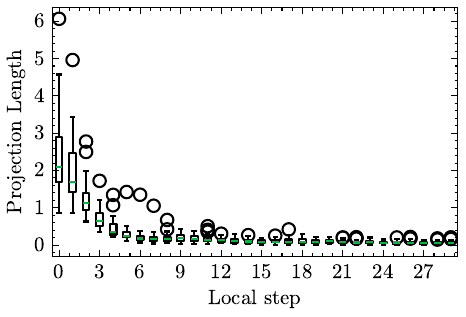}
  \caption{Box plot of the active clients gradient projection length.}
  \label{fig:cs-proj-box}
\end{figure}

As shown in \cref{fig:cs-box}, the average gradient cosine similarity from the mean gradient decreases as the local training goes on.
This indicates the divergence of the gradients among the clients in the later local training.
Furthermore, \cref{fig:cs-proj-box} shows that the average gradient projection on the mean gradient are calculated and gathered across global iterations.
It is clear that the average gradient projection decreases during each local training.
This projection length represents the usefulness of the gradients in the local training.
This confirms the randomness of the gradients in the later local training.

Aggregating the biased momentum in the server side and broadcasting it to the clients gives a suboptimal initialization for the local SGDM optimizer.
Therefore, we need to find a better way to calculate the estimated momentum used in local initialization.

The design of FL are a trade-off between the communication cost and the local training cost.
Ideally, increasing the number of local epochs should reduce the communication cost.
However, the suboptimal momentum initialization makes long local training less effective.

\subsection{Proposed Solution}
The cumulative momentum use in \cref{eq:momentum} assigns more weights to the recent gradients than the past gradients.
This weight strategy suits the centralized training well but not the FL system with heterogeneity.
Gathering such gradients into momentum thus should be rethought.

In order to mitigate the larger bias in the later local training, we propose a reverse exponential decay of the gradients.
That is to assign exponentially decayed weights to the gradients with the time going forward.
The estimated momentum can be calculated as:
\begin{equation}
  \hat{v}_t = (1-\beta) v_0 + (1-\beta) \sum_{i=0}^{t-1} \beta^{i} \nabla f(x_i) + \beta^t \nabla f(x_t)
\end{equation}
Where $\hat{v}_t$ is the estimated momentum at local step $t$ and $v_0$ is the initial momentum inherited from the server side.
The weights of the gradients are exponentially decayed with the time going forward, which is on the contrary to the MFL method.

By using a reverse exponential decay of the gradients, biased gradients in the latest local steps account for less in the cumulative momentum.

\section{Experiments}
In this section, we evaluate the proposed method on the three popular benchmark datasets with different heterogeneity levels.
MNIST~\cite{deng2012mnist} is a handwriting digit dataset with 10 classes of size $28\times28$ greyscale images.
The training set contains 6,000 samples for each class and the test set contains 1,000 samples for each class.
CIFAR10~\cite{Krizhevsky2009LearningML} is a more challenge dataset with 10 classes of size $3\times32\times32$ RGB images.
The training set contains 5,000 samples for each class and the test set contains 1,000 samples for each class.
CIFAR100~\cite{Krizhevsky2009LearningML} dataset is just like the CIFAR-10, except it has 100 classes containing 600 images each.

The baseline algorithm are the MFL which adopt a traditional momentum cumulation.
The proposed method is named as Reversed Momentum Federated Learning (RMFL).

\subsection{Experimental Setup}
According to the difficulty of the dataset, three different neural networks are used respectively.
For MNIST dataset, a simple 2-layer fully connected neural network with 128 hidden units is used.
The activation function is ReLU and the output layer is followed by a Softmax function.
Standard LeNet-5~\cite{lecun1998gradient} is used for CIFAR10 datasets except ReLU is used as the activation function instead of Sigmoid.
ResNet20~\cite{DBLP:conf/cvpr/HeZRS16} is used for the most challenge CIFAR100 dataset.

We simulate the data heterogeneity by altering the class distribution of the client datasets.
In detail, the class distribution $\bm{N}$ for each client are sampled from a Dirichlet distribution $\bm{N} \sim \text{Dir}(\alpha)$ parameterized by $\alpha$.
With a larger $\alpha$, the class distribution among the clients are more even.
With a smaller $\alpha$, the class distribution among the clients are more skewed.
For each dataset, $\alpha$ in scope of [1, 0.1, 0.01] are used to simulate the different heterogeneity levels.

\begin{figure}[htbp]
  \centering
  \includegraphics{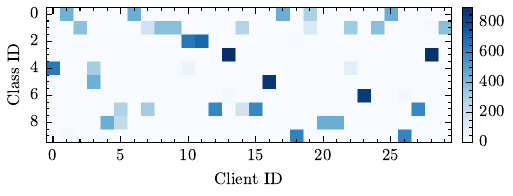}
  \caption{Number of samples for each class of the first 30 clients sampled from Dirichlet distribution with $\alpha=0.01$.}
  \label{fig:dirichlet}
\end{figure}

\Cref{fig:dirichlet} shows the class distribution of the first 30 clients sampled form the Dirichlet distribution with $\alpha=0.01$.
Clearly, the majority of the clients have only data from one or two classes, and it is rare to have data from four or more classes in this setting.

The FL system is composed of 100 clients in which 10 random clients are active in each round.
The basic local training epochs are set to 2, which refers to every local training loops twice over their local dataset.
Larger local training epochs include 5 and 10 are also considered in the experiments investigating the sensitivity to the number of local epochs.

As the MFL parer suggested, we use a fixed momentum coefficient $\beta=0.9$ throughout the experiments.
For the RMFL, we use the same momentum coefficient $\beta=0.9$.
The learning rate for MFL and RMFL are searched in the scope of [0.3, 0.1, 0.03,0.01,0.003,0.001] for the fastest convergence.
For each experiment setting, 10 repeated runs are conducted with random seeds from 0 to 9.
All the results presented are averaged over the 10 runs with standard deviation in parentheses.

\subsection{Results}
Although the test set for MNIST, CIFAR10, and CIFAR100 datasets are balanced, macro F1-score is used alongside with the accuracy as the evaluation metric.
\Cref{tab:results} shows the accuracy and F1-score of the FL models with MFL and RMFL on MNIST, CIFAR10, and CIFAR100 datasets in different heterogeneity levels parameterized by $\alpha$.
It is clear that RMFL outperforms MFL in all the settings for both accuracy and F1-score.
Also, RMFL establishes a smaller standard deviation than MFL in most of the settings, indicating the robustness of the proposed method.

MNIST is relatively easy to learn for both MFL and RMFL, therefore the results are collected in less iterations.
The advantage of RMFL over MFL is more significant in CIFAR10 with increasing heterogeneity from $\alpha=1$ to $\alpha=0.01$.
This is consistent with the intuition that the proposed method is more effective in the more heterogeneous data distribution.
As CIFAR100 contains fewer samples for each class, the gap between RMFL and MFL is not directly proportional to the heterogeneity level.
However, RMFL still outperforms MFL in all heterogeneity levels in CIFAR100 dataset.

\begin{table}[htbp]
  \centering
  \begin{tabular}{lllrr}
    \toprule
    Dataset                      & $\alpha$                 & Met      & MFL           & RMFL                    \\
    \midrule
    \multirow[c]{6}{*}{MNIST}    & \multirow[c]{2}{*}{1}    & Accuracy & 0.888 (0.097) & \bfseries 0.940 (0.014) \\
                                 &                          & F1-score & 0.883 (0.106) & \bfseries 0.939 (0.014) \\
    \cline{2-5}
                                 & \multirow[c]{2}{*}{0.1}  & Accuracy & 0.928 (0.007) & \bfseries 0.929 (0.005) \\
                                 &                          & F1-score & 0.927 (0.008) & \bfseries 0.928 (0.005) \\
    \cline{2-5}
                                 & \multirow[c]{2}{*}{0.01} & Accuracy & 0.871 (0.023) & \bfseries 0.920 (0.017) \\
                                 &                          & F1-score & 0.868 (0.024) & \bfseries 0.919 (0.018) \\
    \midrule
    \multirow[c]{6}{*}{CIFAR10}  & \multirow[c]{2}{*}{1}    & Accuracy & 0.527 (0.015) & \bfseries 0.555 (0.014) \\
                                 &                          & F1-score & 0.517 (0.017) & \bfseries 0.551 (0.015) \\
    \cline{2-5}
                                 & \multirow[c]{2}{*}{0.1}  & Accuracy & 0.435 (0.012) & \bfseries 0.506 (0.023) \\
                                 &                          & F1-score & 0.411 (0.015) & \bfseries 0.495 (0.026) \\
    \cline{2-5}
                                 & \multirow[c]{2}{*}{0.01} & Accuracy & 0.247 (0.057) & \bfseries 0.472 (0.019) \\
                                 &                          & F1-score & 0.183 (0.065) & \bfseries 0.452 (0.021) \\
    \midrule
    \multirow[c]{6}{*}{CIFAR100} & \multirow[c]{2}{*}{1}    & Accuracy & 0.318 (0.011) & \bfseries 0.384 (0.011) \\
                                 &                          & F1-score & 0.314 (0.012) & \bfseries 0.383 (0.011) \\
    \cline{2-5}
                                 & \multirow[c]{2}{*}{0.1}  & Accuracy & 0.243 (0.008) & \bfseries 0.286 (0.014) \\
                                 &                          & F1-score & 0.233 (0.009) & \bfseries 0.275 (0.015) \\
    \cline{2-5}
                                 & \multirow[c]{2}{*}{0.01} & Accuracy & 0.169 (0.010) & \bfseries 0.195 (0.009) \\
                                 &                          & F1-score & 0.150 (0.011) & \bfseries 0.179 (0.010) \\
    \bottomrule\\
  \end{tabular}

  \caption{Accuracy and F1-score of FL models with MFL and RMFL on MNIST and CIFAR10 datasets in different heterogeneity level.
    The results are averaged over 5 runs with standard deviation in parentheses.
    The best results are highlighted in bold for each setting.
  }
  \label{tab:results}
\end{table}

\begin{figure}[htbp]
  \centering
  \includegraphics{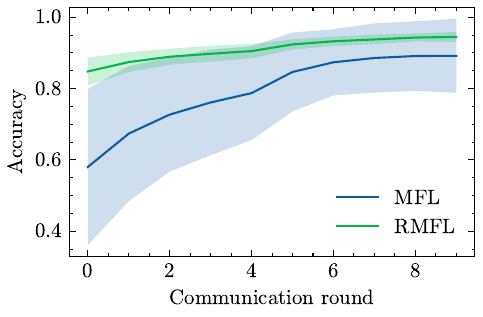}
  \caption{Accuracy of FL models with MFL and RMFL on MNIST dataset with $\alpha=1$}
  \label{fig:mnist-1-acc}
\end{figure}

\begin{figure}[htbp]
  \centering
  \includegraphics{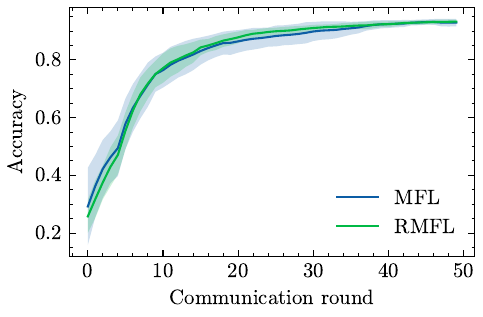}
  \caption{Accuracy of FL models with MFL and RMFL on MNIST dataset with $\alpha=0.1$}
  \label{fig:mnist-0.1-acc}
\end{figure}

\begin{figure}[htbp]
  \centering
  \includegraphics{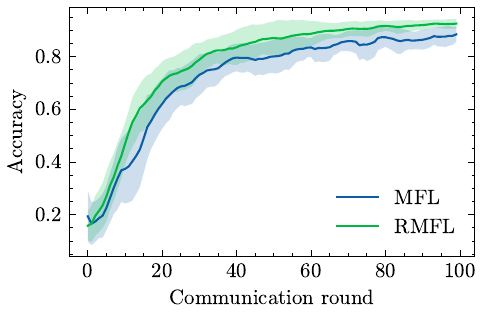}
  \caption{Accuracy of FL models with MFL and RMFL on MNIST dataset with $\alpha=0.01$}
  \label{fig:mnist-0.01-acc}
\end{figure}

The accuracy curve of MFL and RMFL on MNIST dataset with $\alpha=1$, $\alpha=0.1$, and $\alpha=0.01$ are shown in \cref{fig:mnist-1-acc,fig:mnist-0.1-acc,fig:mnist-0.01-acc} respectively.
Although MNIST is easy task across selected heterogeneity levels, RMFL still shows similar or better performance than MFL.
Especially with $\alpha=0.01$, RMFL converges faster and achieves better accuracy than MFL.

\begin{figure}[htbp]
  \centering
  \includegraphics{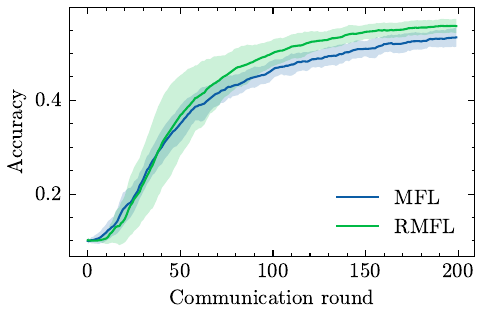}
  \caption{Accuracy of FL models with MFL and RMFL on CIFAR10 dataset with $\alpha=1$}
  \label{fig:cifar10-1-acc}
\end{figure}

\begin{figure}[htbp]
  \centering
  \includegraphics{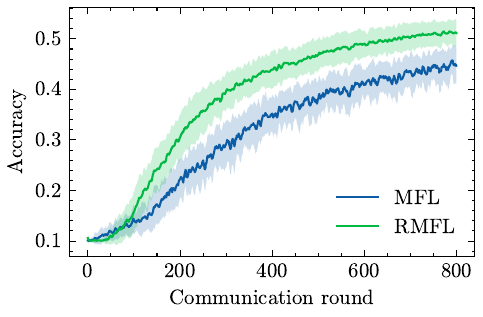}
  \caption{Accuracy of FL models with MFL and RMFL on CIFAR10 dataset with $\alpha=0.1$}
  \label{fig:cifar10-0.1-acc}
\end{figure}

\begin{figure}[htbp]
  \centering
  \includegraphics{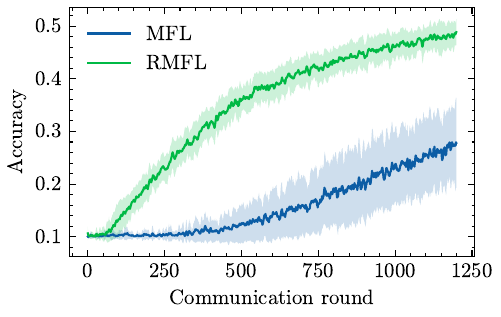}
  \caption{Accuracy of FL models with MFL and RMFL on CIFAR10 dataset with $\alpha=0.01$}
  \label{fig:cifar10-0.01-acc}
\end{figure}

\Cref{fig:cifar10-1-acc,fig:cifar10-0.1-acc,fig:cifar10-0.01-acc} show the accuracy curve of MFL and RMFL on CIFAR10 dataset with $\alpha=1$, $\alpha=0.1$, and $\alpha=0.01$ respectively.
With $\alpha=1$, RMFL shows similar performance as MFL in the initial 50 global iterations.
This indicates the clients gradients are less biased in this stage.
After 50 global iterations, RMFL outperforms MFL and converges faster as the increasing bias in the gradients.
Similar phenomena are observed in the CIFAR10 dataset with $\alpha=0.1$ before and after 100 global iterations.
And RMFL keeps 20\% better accuracy than MFL after that.
With $\alpha=0.01$, RMFL shows significant advantage over MFL in the whole training process.
This could be the bias in the gradients severely affect the convergence of MFL, while RMFL is more capable of handling the biased gradients.

\begin{figure}[htbp]
  \centering
  \includegraphics{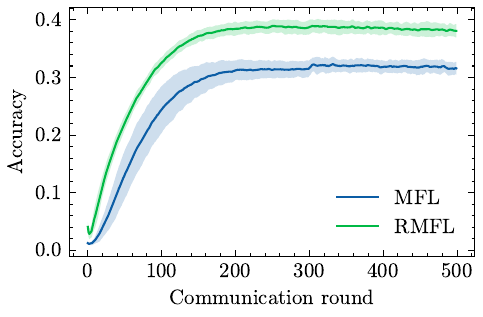}
  \caption{Accuracy of FL models with MFL and RMFL on CIFAR100 dataset with $\alpha=1$}
  \label{fig:cifar100-1-acc}
\end{figure}

For CIFAR100 dataset, as there is more classes and fewer samples for each class, the heterogeneity is more severe with same $\alpha$.
\Cref{fig:cifar100-1-acc} shows the accuracy curve of MFL and RMFL on CIFAR100 dataset with $\alpha=1$, in which RMFL outperforms MFL in the whole training process for around 30\%.
With $\alpha=0.1$ and $\alpha=0.01$, RMFL still shows better performance than MFL after initial training stage.

\begin{figure}[htbp]
  \centering
  \includegraphics{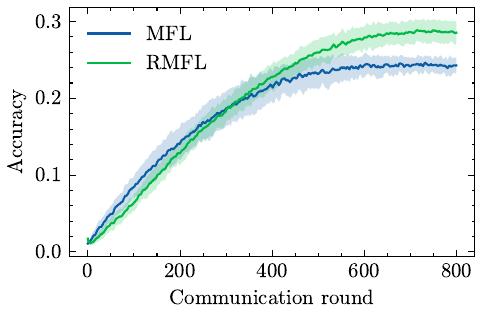}
  \caption{Accuracy of FL models with MFL and RMFL on CIFAR100 dataset with $\alpha=0.1$}
  \label{fig:cifar100-0.1-acc}
\end{figure}

\begin{figure}[htbp]
  \centering
  \includegraphics{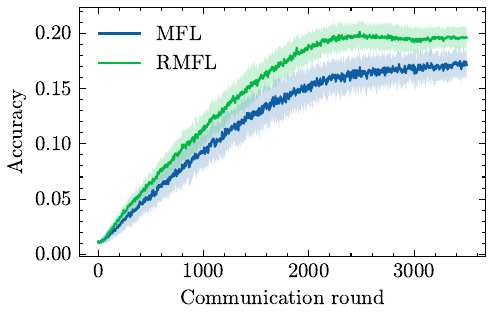}
  \caption{Accuracy of FL models with MFL and RMFL on CIFAR100 dataset with $\alpha=0.01$}
  \label{fig:cifar100-0.01-acc}
\end{figure}

\subsection{Sensitivity to the Number of Local Epochs}
FL algorithms are sensitivity to the number of local epochs.
As the number of local epochs increases, the similarity between the local updates decreases.
This becomes more challenge in heterogeneous data distribution.

MFL are also vulnerable to increasing number of local epochs.
As the cumulative momentum are less accurate due to more biased gradients involved, the initialization of the next round of local training are less effective.
In addition to previous experiments with 2 local epochs, we also test both algorithms with 5 and 10 local epochs.

\begin{figure}[htbp]
  \centering
  \includegraphics{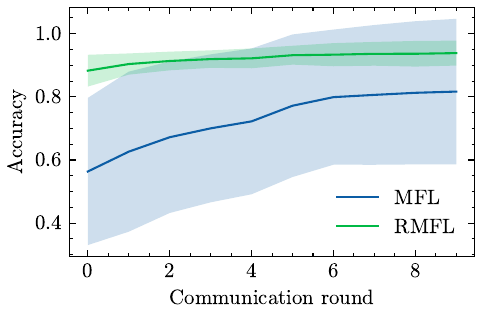}
  \caption{Accuracy of FL models with MFL and RMFL on MNIST dataset with $\alpha=1$ and 5 local epochs}
  \label{fig:mnist-1-acc-5}
\end{figure}

\begin{figure}[htbp]
  \centering
  \includegraphics{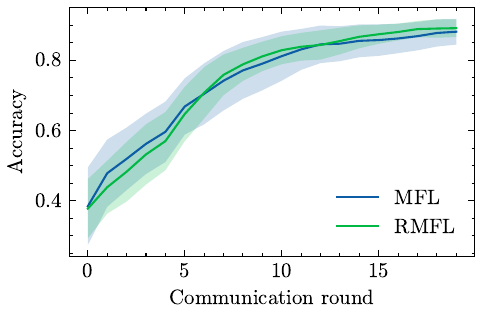}
  \caption{Accuracy of FL models with MFL and RMFL on MNIST dataset with $\alpha=0.1$ and 5 local epochs}
  \label{fig:mnist-0.1-acc-5}
\end{figure}

\begin{figure}[htbp]
  \centering
  \includegraphics{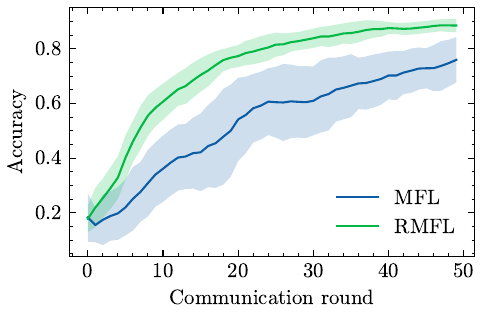}
  \caption{Accuracy of FL models with MFL and RMFL on MNIST dataset with $\alpha=0.01$ and 5 local epochs}
  \label{fig:mnist-0.01-acc-5}
\end{figure}

\begin{figure}[htbp]
  \centering
  \includegraphics{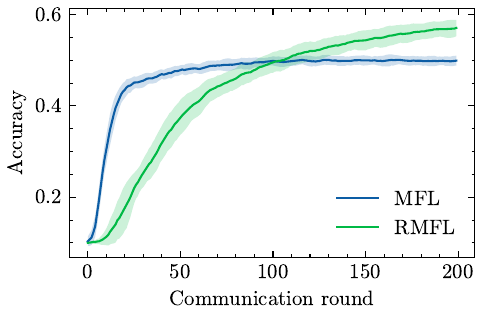}
  \caption{Accuracy of FL models with MFL and RMFL on CIFAR10 dataset with $\alpha=1$ and 5 local epochs}
  \label{fig:cifar10-1-acc-5}
\end{figure}

\begin{figure}[htbp]
  \centering
  \includegraphics{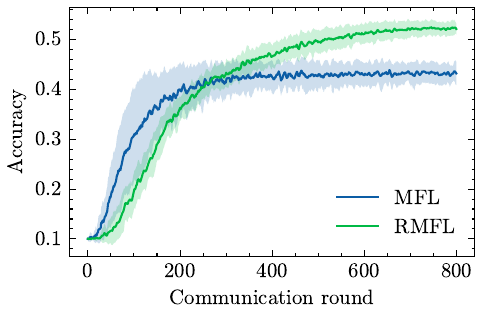}
  \caption{Accuracy of FL models with MFL and RMFL on CIFAR10 dataset with $\alpha=0.1$ and 5 local epochs}
  \label{fig:cifar10-0.1-acc-5}
\end{figure}

\begin{figure}[htbp]
  \centering
  \includegraphics{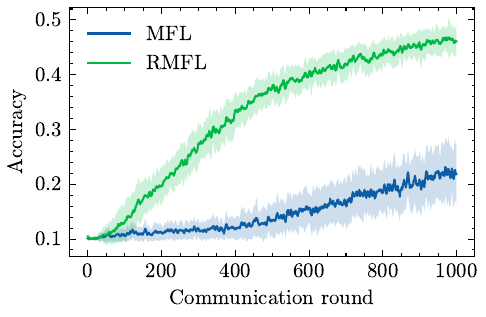}
  \caption{Accuracy of FL models with MFL and RMFL on CIFAR10 dataset with $\alpha=0.01$ and 5 local epochs}
  \label{fig:cifar10-0.01-acc-5}
\end{figure}

\Cref{fig:mnist-1-acc-5,fig:mnist-0.1-acc-5,fig:mnist-0.01-acc-5} show the accuracy curve of MFL and RMFL on MNIST dataset with 5 local epochs and  $\alpha=\{1,0.1,0.01\}$ respectively.
RMFL clearly outperforms MFL in all the settings.
Similar results are observed in CIFAR10 dataset with 5 local epochs and $\alpha=\{1,0.1,0.01\}$ as shown in \cref{fig:cifar10-1-acc-5,fig:cifar10-0.1-acc-5,fig:cifar10-0.01-acc-5}.

\begin{figure}[htbp]
  \centering
  \includegraphics{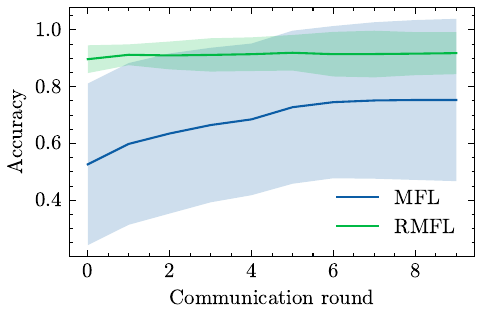}
  \caption{Accuracy of FL models with MFL and RMFL on MNIST dataset with $\alpha=1$ and 10 local epochs}
  \label{fig:mnist-1-acc-10}
\end{figure}

\begin{figure}[htbp]
  \centering
  \includegraphics{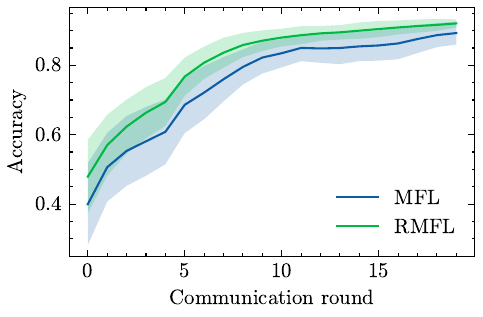}
  \caption{Accuracy of FL models with MFL and RMFL on MNIST dataset with $\alpha=0.1$ and 10 local epochs}
  \label{fig:mnist-0.1-acc-10}
\end{figure}

\begin{figure}[htbp]
  \centering
  \includegraphics{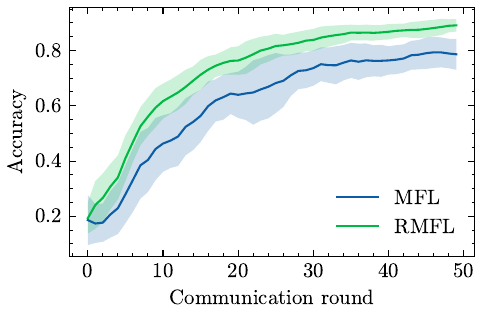}
  \caption{Accuracy of FL models with MFL and RMFL on MNIST dataset with $\alpha=0.01$ and 10 local epochs}
  \label{fig:mnist-0.01-acc-10}
\end{figure}

Further, increasing the number of local epochs to 10, RMFL still outperforms MFL in all the settings.
\Cref{fig:mnist-1-acc-10,fig:mnist-0.1-acc-10,fig:mnist-0.01-acc-10} show the accuracy curve of MFL and RMFL on MNIST dataset with 10 local epochs and  $\alpha=\{1,0.1,0.01\}$ respectively.

\begin{figure}[htbp]
  \centering
  \includegraphics{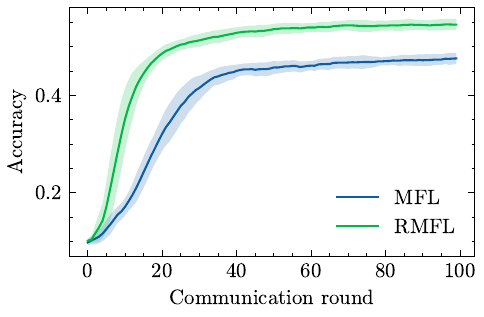}
  \caption{Accuracy of FL models with MFL and RMFL on CIFAR10 dataset with $\alpha=1$ and 10 local epochs}
  \label{fig:cifar10-1-acc-10}
\end{figure}

\begin{figure}[htbp]
  \centering
  \includegraphics{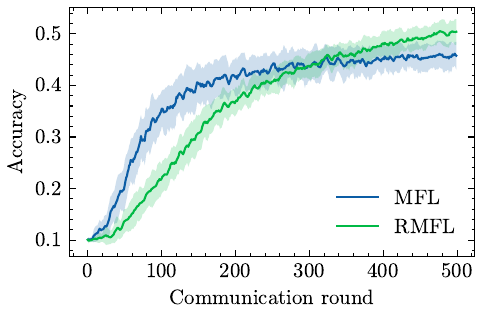}
  \caption{Accuracy of FL models with MFL and RMFL on CIFAR10 dataset with $\alpha=0.1$ and 10 local epochs}
  \label{fig:cifar10-0.1-acc-10}
\end{figure}

\begin{figure}[htbp]
  \centering
  \includegraphics{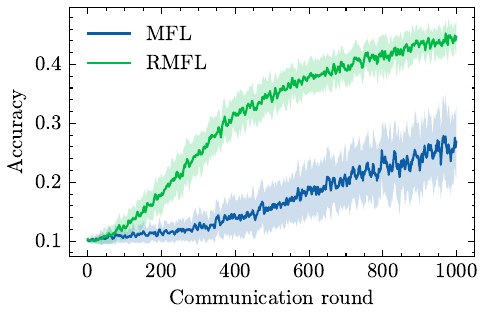}
  \caption{Accuracy of FL models with MFL and RMFL on CIFAR10 dataset with $\alpha=0.01$ and 10 local epochs}
  \label{fig:cifar10-0.01-acc-10}
\end{figure}

RMFL even establishes a larger gap over MFL with 10 local epochs.
\Cref{fig:cifar10-1-acc-10,fig:cifar10-0.1-acc-10,fig:cifar10-0.01-acc-10} show the accuracy curve of MFL and RMFL on CIFAR10 dataset with 10 local epochs and  $\alpha=\{1,0.1,0.01\}$ respectively.

\section{Conclusion}
In this chapter, we investigate the advanced optimization techniques in FL systems.
Specifically, we focus on the momentum based optimization techniques on the local training stage.
The findings show that the traditional momentum cumulation in centralized training is suboptimal in FL systems with heterogeneous data distribution.
The exponential moving average of the gradients in the momentum cumulation assigns more weights to the recent gradients than the past gradients.
This works against the gradient divergence among the clients in FL as the recent gradients are biased due to the heterogeneity.
We propose a reverse exponential decay of the gradients in the momentum cumulation to mitigate the bias in the gradients.
By assigning more weights to the early gradients than the recent gradients locally, the proposed method is more capable of handling the biased gradients in the later local training.
The experiments on MNIST, CIFAR10, and CIFAR100 datasets with different heterogeneity levels show that the proposed method outperforms the traditional momentum cumulation in all the settings.
The advanced optimization techniques in FL systems evolves rapidly.
There may exist even better cumulation approaches to handle the biased gradients in the heterogeneous FL systems.
There are other algorithms utilizing the momentum differently in the local training stage.
The idea of correctly initialize them may also apply to these algorithms.

\bibliographystyle{IEEEtran}
\bibliography{references}

\end{document}